\documentclass[runningheads]{llncs}
\usepackage{tikz}
\usepackage{float}
\usepackage{tabularray}
\usepackage{graphicx} % Required for including images
\usepackage{subcaption} % Required for subfigures
\usetikzlibrary{shapes,arrows,positioning}
\usetikzlibrary{shapes.geometric, arrows, positioning}
\tikzstyle{startstop} = [rectangle, rounded corners, minimum width=3cm, minimum height=1cm,text centered, draw=black, fill=red!30]
\tikzstyle{process} = [rectangle, minimum width=3cm, minimum height=1cm, text centered, draw=black, fill=orange!30]
\tikzstyle{arrow} = [thick,->,>=stealth]
\usepackage{booktabs} % For prettier tables
\usepackage{tabularx} % For adjustable width tables
\usepackage[T1]{fontenc}
\usepackage{graphicx}
\begin{document}
\title{Learning Paradigms and Modelling Methodologies for Digital Twins in Process Industry}

\author{Michael Mayr\inst{1} \and
Georgios C. Chasparis\inst{1} \and Josef Küng\inst{2}
}

\authorrunning{Michael Mayr, et al.}
\institute{
Software Competence Center Hagenberg, Softwarepark 32a, 4232 Hagenberg, Austria \\ 
\email{\{michael.mayr,georgios.chasparis\}@scch.at}
\and
Institute for Application-Oriented Knowledge Processing, Johannes Kepler University, Altenberger Str. 69, 4040 Linz, Austria \\ 
\email{jkueng@faw.jku.at}
}
\maketitle              % typeset the header of the contribution
\begin{abstract}
Central to the digital transformation of the process industry are Digital Twins (DTs), virtual replicas of physical manufacturing systems that combine sensor data with sophisticated data-based or physics-based models, or a combination thereof, to tackle a variety of industrial-relevant tasks like process monitoring, predictive control or decision support. 
The backbone of a DT, i.e. the concrete modelling methodologies and architectural frameworks supporting these models, are complex, diverse and evolve fast, necessitating a thorough understanding of the latest state-of-the-art methods and trends to stay on top of a highly competitive market. From a research perspective, despite the high research interest in reviewing various aspects of DTs, structured literature reports specifically focusing on unravelling the utilized learning paradigms (e.g. self-supervised learning) for DT-creation in the process industry are a novel contribution in this field.
This study aims to address these gaps by (1) systematically analyzing the modelling methodologies (e.g. Convolutional Neural Network, Encoder-Decoder, Hidden Markov Model) and paradigms (e.g. data-driven, physics-based, hybrid) used for DT-creation; (2) assessing the utilized learning strategies (e.g. supervised, unsupervised, self-supervised); (3) analyzing the type of modelling task (e.g. regression, classification, clustering); and (4) identifying the challenges and research gaps, as well as, discuss potential resolutions provided.


\end{abstract}

\section{Introduction and Motivation}

The number of sensors and the corresponding data produced in the process industry continuously increases. This uptrend is part of the Industry 4.0 revolution, enabling a rich source of sensor data \cite{lasi_industry_2014}, which presents an unprecedented opportunity to harness complex sensor data for enhancing operational efficiency. The rapidly evolving market demands and the necessity for quick decision-making introduce considerable challenges to industrial manufacturers \cite{zeng_virtual_2017}. Furthermore, manufacturers within the European Union are obligated to comply with energy efficiency standards \cite{EU2050}, aiming to neutralize the carbon footprint of the production facilities till 2050. An intelligently managed and regulated production is essential to remain competitive and meet efficiency goals. \\
The strategy of employing Digital Twins is recognized as a key enabler for this digital shift, aiming to increase competitiveness, productivity, and efficiency \cite{kritzinger_digital_2018}. Rasheed et al. \cite{rasheed_digital_2020} define DT as a "virtual representation of a physical asset or process enabled through data and simulators for real-time prediction, optimization, monitoring, controlling, and improved decision-making". Recently, the evolution of DTs embraced cognitive capabilities, introducing Adaptive, Intelligent, and Cognitive DTs (e.g.  \cite{abburu_cognitwin_2020}), showcasing their progression towards autonomy and intelligence. \\
Tracking concrete modelling methodologies, learning paradigms, and architectural designs is especially important for the latest concepts, i.e., cognitive or intelligent DTs, to help researchers and industrial practitioners adopt them more efficiently. Cognitive DTs aim to achieve elements of cognition, e.g. perception (i.e. abstracting meaningful data representations for subsequent processing), attention (i.e. focusing by intent or driven by signals on specific tasks, goals or data, e.g. focusing on certain aspects of the multi-dimensional and multi-modal high-volume and high-velocity data produced by Industrial Internet of Things (IIoT) devices), memory (i.e. working memory, episodic memory and semantic memory), learning (i.e. transforming insights from the physical Twin into generalizable knowledge for unseen scenarios), or reasoning \cite{mortlock_graph_2022}. 
Among the learning paradigms of machine learning models, transfer learning and self-supervised learning stand out as promising research directions for enabling cognitive DTs in the process industry. Transfer learning (TL) leverages knowledge, i.e. general representations, learned from pre-training on large-scale datasets and applies it to a target task with limited labelled data. Self-supervised learning does not rely on labelled data but instead learns the general representations from pre-text tasks like masked imputation, commonly using encoder-decoder-related structures (see \cite{yan_toward_2023}). Both learning paradigms have shown tremendous success in the natural language processing domain, where most of the success traces back to the paper of Vaswani et al. \cite{vaswani_attention_2017}. The used learning paradigm and modelling methodology are promising research directions for DTs in general, but specifically for the process industry since the data accumulated in such industries is commonly of high volume, variety, variability and veracity, making manual labelling for specific use cases (e.g. detecting anomalies, predicting key performance indicators, classifying process states, simulating process behaviour based on different control inputs) a time-consuming task for experts and researchers, or even impossible if the modelled process does not allow for exploring the behaviour out-of-domain, i.e. experiments outside of the "normal", business-critical operation.

\subsection{Research Questions (RQs)} \label{researchquestions}
Guided by the preliminary findings of this review, we propose several research questions aimed at furthering the understanding of modelling methodologies and learning paradigms of DTs in the industrial sector:

\begin{itemize}
    \item \textbf{RQ 1:} What are the state-of-the-art modelling methodologies, and how has their usage evolved? Are Encoder-Decoder architectures on the rise, possibly signalling the adoption of transformer-like architectures to the industrial DT domain?
    \item \textbf{RQ 2:} What are the commonly utilized learning paradigms (i.e. unsupervised, supervised, self-supervised, transfer-learning, etc.)? What is the distribution of data-driven and hybrid modelling approaches? Is the self-supervised learning paradigm for generating general and transferable knowledge already explored in the industrial DT domain?
    \item \textbf{RQ 3:} Are DT research and application studies more focused on the evaluation of classification, clustering tasks (e.g. anomaly detection) or regression tasks (e.g. forecasting, imputation, etc.)?
\end{itemize} 

\subsection{Structure of Review} 
\label{sec:structure}
This review is organized into the following sections to systematically address the mentioned research questions, starting with the search strategy description, followed by the review report and detailed analysis of the selected primary studies. By structuring the review in this manner, we aim to provide a comprehensive overview of the current state of digital twin modelling methodologies and learning paradigms in the process industry, highlighting innovative practices and future opportunities for research and application.

\begin{itemize}
    \item \textbf{Sec.~\ref{sec:searchstrategy} - Search Strategy:} Details the methodology used to select and analyze relevant literature, including data sources, selection criteria, and the procedure for selecting primary studies.
    \item \textbf{Sec.~\ref{sec:reportreview} - Reporting the Review:} Presents an overview of the studies included in the review, featuring publication trends and further synthesis modelling methodologies and learning paradigms based on selected primary studies. 
    \item \textbf{Sec.~\ref{sec:evaluatereview} - Evaluating RQs on Primary Studies}: Offers a detailed analysis of the primary studies, focusing on their contributions to modelling methodologies, learning paradigms, and the tackled tasks in the context of the process industry. This section also discusses the challenges identified and potential solutions.
    \item \textbf{Sec.~\ref{sec:discussionandconclusion} - Discussion and Conclusion}: Synthesizes the review findings, discusses the implications for industry and academia, identifies current literature gaps, and proposes future research directions.
\end{itemize}

\section{Literature Search Strategy} 
\label{sec:searchstrategy}
This review systematically searches academic and industry literature across several well-known databases, including IEEE Xplore\footnote[1]{\url{https://ieeexplore.ieee.org/}}, Scopus (Elsevier)\footnote[2]{\url{https://scopus.com/}}, and ScienceDirect\footnote[3]{\url{https://sciencedirect.com/}}. The search captures a broad spectrum of research focusing on DTs in the process industry w.r.t learning paradigms and modelling methodologies. The search string (see Tab.~\ref{tab:screening_criteria}) captures all relevant papers related to DTs that mention modelling paradigms in the abstract, keywords or publication title. The selection criteria (see Tab.~\ref{tab:selection_criteria}) ensure the selection of studies that are directly relevant to the objectives of this review.

\begin{table}[htbp]
    \centering
    \caption{Screening criteria for literature search}
    \label{tab:screening_criteria}
    \begin{tabularx}{\textwidth}{lX}
        \toprule
        \textbf{Database} & Scopus, IEEE Xplore, ScienceDirect \\
        \midrule
        \textbf{Search string} & ("digital twin*" OR "cognitive twin*" OR "cyber twin*" OR "adaptive twin*" OR "intelligent twin*") AND ("unsupervised" OR "supervised" OR "semi-supervised" OR "self-supervised") \\
        \midrule
        \textbf{Document type} & Journal and conference papers \\
        \bottomrule
    \end{tabularx}
\end{table}

\begin{table}[htbp]
    \centering
    \caption{Studies selection criteria}
    \label{tab:selection_criteria}
    \begin{tabularx}{\textwidth}{lX}
        \toprule
        \multicolumn{2}{c}{\textbf{Inclusion Criteria}} \\
        \midrule
        I-1 & Publications published as full research in relevant conferences and journals. \\
        I-2 & Scholarly literature, including books, book chapters, and technical reports. \\
        I-3 & Research focused on digital twins (DT) within the process industry. \\
        I-4 & Studies using machine learning, deep learning, other data-driven techniques, physics-based techniques, or a combination in the context of digital twins. \\
        I-5 & Contributions significantly advancing theory, methodology, or application of digital twins in the process industry. \\
        \midrule
        \multicolumn{2}{c}{\textbf{Exclusion Criteria}} \\
        \midrule
        E-1 & Non-full-length research articles, including abstracts, essays. \\
        E-2 & Publications lacking an accessible abstract or missing metadata. \\
        E-3 & Documents not written in English. \\
        E-4 & Studies on digital twins that omit explicit discussion of the modelling methodology and learning paradigm as well as tackled tasks. \\
        E-5 & Research does not specifically address the application of DTs in process industry. \\
        \bottomrule
    \end{tabularx}
\end{table}

\subsection{Quality Assessment Checks}
Each research left after filtering based on the selection criteria (see Tab.\ref{tab:selection_criteria}) based on the abstract, keywords and title, gets checked on specific aspects of quality measurements in the full-text of the research, which are outlined in Tab.\ref{tab:quality_assessment}. \textbf{A score of 2} is awarded if the quality item is fully met, demonstrating comprehensive adherence to the criteria. \textbf{A score of 1} is given if the item is partially met, indicating that while there is some adherence to the criteria, certain aspects are lacking or not fully developed.\textbf{A score of 0} is assigned if the criterion is not met or the information is unavailable, reflecting a complete absence of the required information or a failure to meet the specified quality standard.

\begin{table}
\centering
\caption{Quality Assessment List}
\label{tab:quality_assessment}
\begin{tabularx}{\textwidth}{>{\bfseries}l|X} % Makes the first column bold
\toprule
Quality Item & Description \\
\midrule
Problem & Defines the research problem and objectives. \\
Context & Describes the research's industrial or practical context. \\
Methodology & Details the concrete modelling methodology. \\
Learning & Details the learning paradigm. \\
Task & Details the modelling task. \\
Architecture & Explains the architectural components. \\
Experiments & Evaluates on real-world or synthetic data. \\
Limitation & Discusses limitations and future research directions. \\
\bottomrule
\end{tabularx}
\end{table}

\subsection{Selection of Primary Studies}
Each research left after filtering based on the selection criteria (see Tab.\ref{tab:selection_criteria}) is included as a primary study. All full texts, if available, of primary studies are scanned and evaluated based on the quality assessment list (see Tab.\ref{tab:quality_assessment}). This process ensures the fine-grained inclusion of research items that provide significant insights into modelling, learning, and architectural-related aspects of DT-creation for the process industry. 

\subsection{Data Synthesis and Analysis Approach}
Data synthesis constitutes gathering, summarizing, and interpreting the data extracted from the primary studies. It is a crucial phase where qualitative and quantitative statistics are drawn and further analyzed. The approach focuses on synthesizing data specifically to address the pre-defined research questions (see Sec.~\ref{researchquestions}). We will extract each primary study's data items specified in Tab.~\ref{tab:data_items}. This extracted data will be collectively analyzed to synthesize a comprehensive overview addressing each research question. For instance, we will gather and analyze all specific modelling methodologies mentioned in the studies (RQ 1) and subsequently visualize the findings. A similar approach will be applied to the other research questions, enabling us to identify patterns, trends, and areas lacking in the current literature.

\begin{table}[htbp]
    \centering
    \caption{Data Items Extracted from Primary Studies}
    \begin{tabularx}{\textwidth}{lXl} % Adjust column width automatically
    \toprule
    \textbf{Data Item} & \textbf{Description} & \textbf{Relevant RQ} \\
    \midrule
    DOI & Unique identifier for research & Documentation \\
    Title & Title of the study & Documentation \\
    Keywords & Indexed Keywords & Demographics \\
    Year & Year of publication & Demographics \\
    Modelling Method & Modelling methodology used, e.g., CNN, AE, HMM, etc. & RQ 1 \\
    Learning Paradigm & Learning paradigm used, e.g., supervised, semi-supervised, unsupervised, self-supervised, etc. & RQ 2 \\
    Modelling Type & Modelling type used, e.g., data-based, physics-based, hybrid, etc. & RQ 2 \\
    Modelling Task & Model is evaluated or developed as a, e.g., regression-related model, classification-related model, clustering-related model & RQ 3 \\
    Architecture & Technology stacks, e.g., frameworks, concepts or stacks proposed to host model(s) & Documentation \\
    Use-Case & Applied use-case, e.g. anomaly detection & Documentation \\
    \bottomrule
    \end{tabularx}
    \label{tab:data_items}
\end{table}

\section{Reporting the Review}
\label{sec:reportreview}
In this section, we report the intermediate steps of the selection process, give an overview of all the studies selected by the selection criteria, and give an overview of the primary studies after scanning the full texts.

\begin{table}[htbp]
    \centering
    \caption{Study Selection Process. }
    \begin{tabularx}{\textwidth}{Xrr} % Adjust column width automatically
    \toprule
    \textbf{Selection Stage} & \textbf{Number of Studies} \\
    \midrule
    Studies matching the search query & 326 & \\
    Studies after filtering (see Tab.\ref{tab:selection_criteria}) & 40 \\
    Studies after reference snowballing & 43 \\
    Studies after full-text review (see Tab.\ref{tab:quality_assessment}) & 31 \\
    \bottomrule
    \end{tabularx}
    \label{tab:study_selection}
\end{table}

\subsection{Overview of All Studies}
\begin{figure}[htbp]
\includegraphics[width=0.49\textwidth]{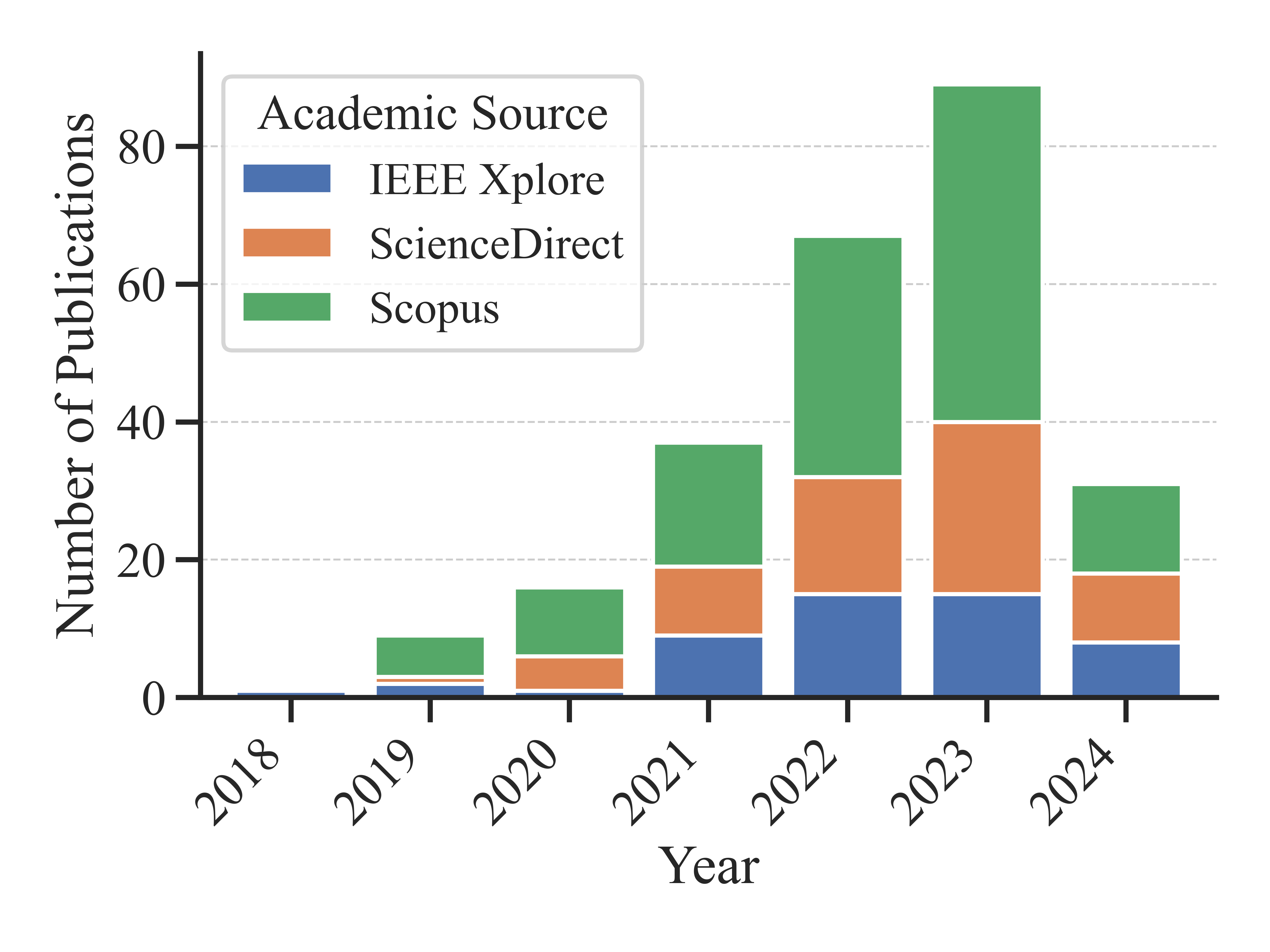}
\includegraphics[width=0.49\textwidth]{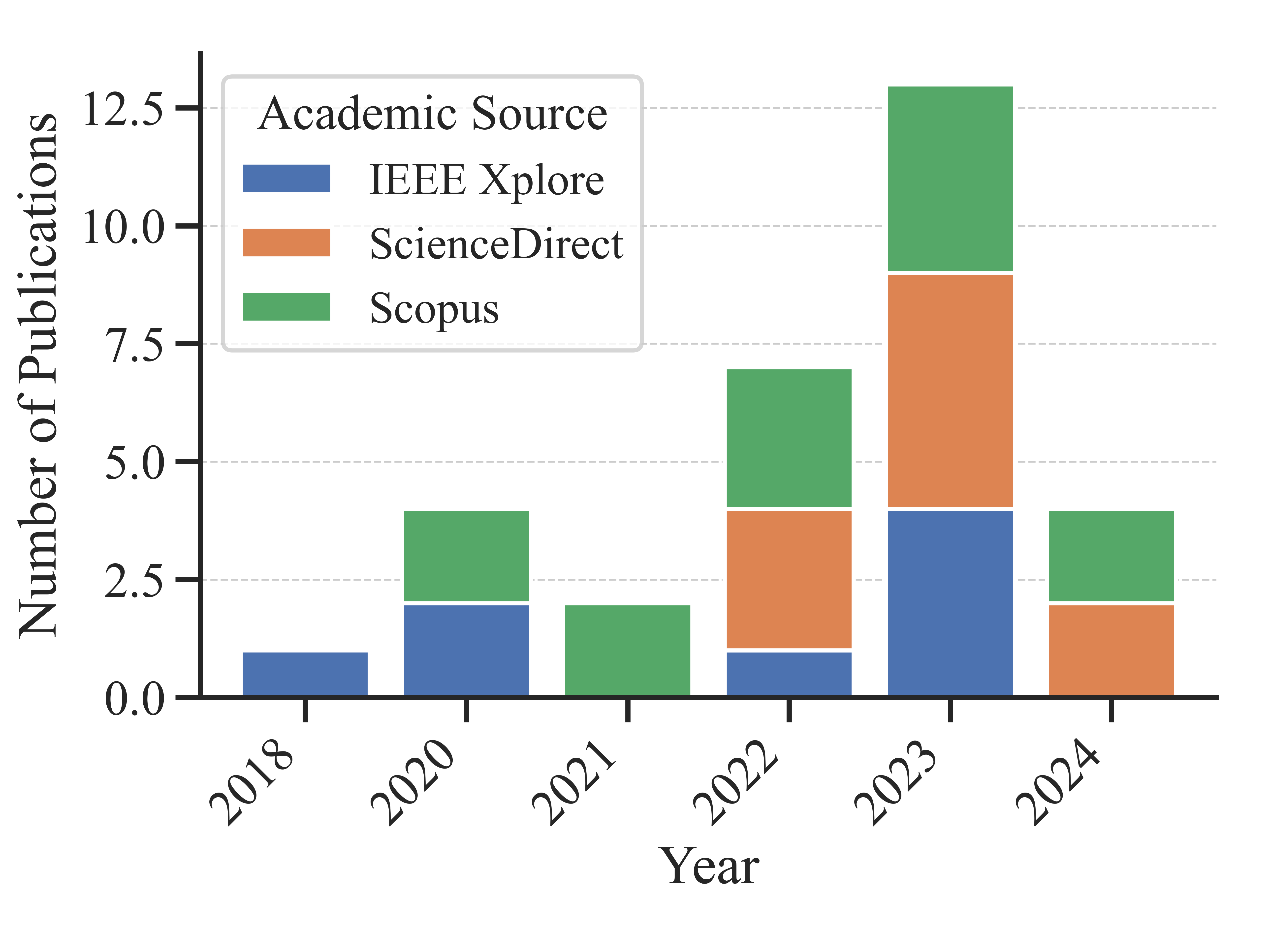}
\caption{(Left) The number of publications matching the search query over time. (Right) The number of publications matching the search query and passing the selection criteria over time.} \label{fig
}
\label{fig:numberofpublicationsallstudies}
\end{figure}

This section presents an aggregate analysis of the studies identified through the search strategy. Critical trends over the years are highlighted in Fig.\ref{fig:numberofpublicationsallstudies}, illustrating the growing interest and evolution of DT-related modelling in the industrial sector.

\subsection{Overview of All Primary Studies}
\begin{figure}[htbp!]
\centering
\includegraphics[width=0.75\textwidth]{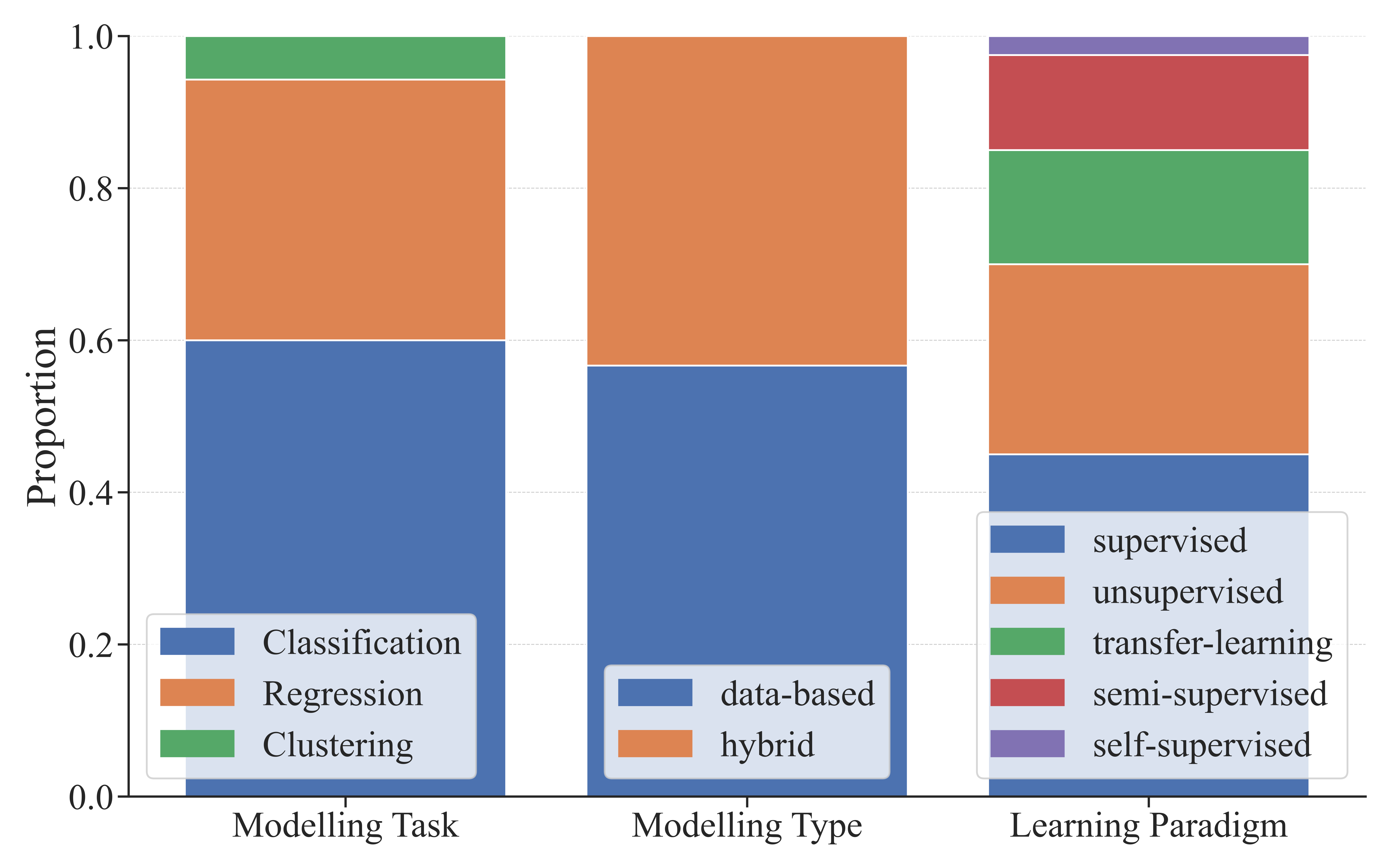}
\caption{Proportions of tasks, types and learning paradigms for the analyzed primary studies.} \label{fig:tasktypelearningboxplot}
\end{figure}

\begin{figure}[htbp!]
\centering
\includegraphics[width=0.75\textwidth]{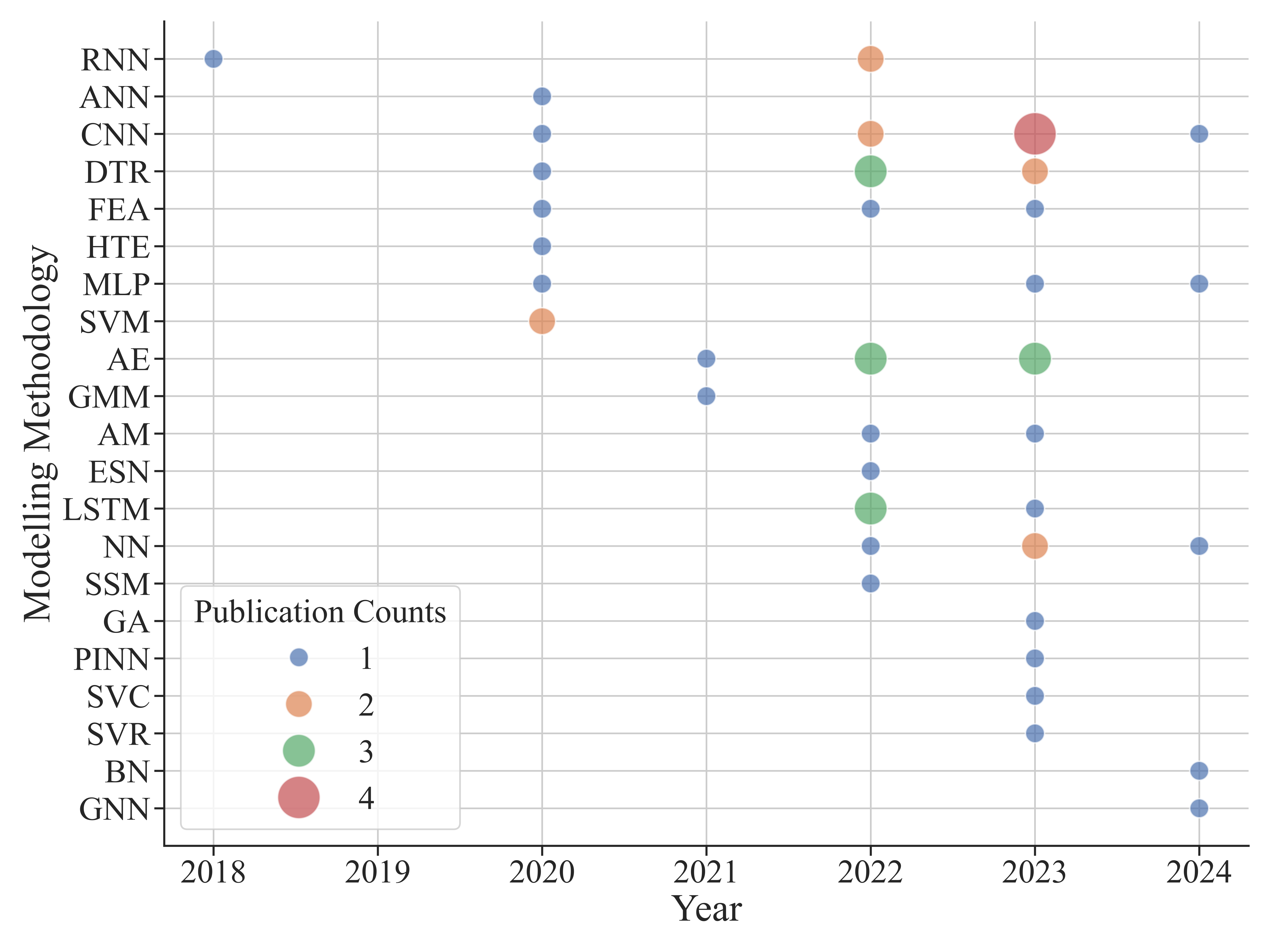}
\caption{Temporal development of the different modelling methodologies for DTs.} \label{fig:TopMMDescOverYear_ScatterBubble_Legend}
\end{figure}

The primary studies, selected based on their comprehensive coverage of digital twin technologies and their impact on industrial manufacturing, are examined in greater detail. This includes a deeper dive into the modelling methodologies, learning paradigms and architectural components discussed across these studies. The number of selected primary studies for each year divided into the different data sources is denoted in Fig.\ref{fig:numberofpublicationsallstudies} on the right-hand side. In addition, the proportions of the modelling task, the modelling type of the DT, and the utilized learning paradigm are visualized in Fig.\ref{fig:tasktypelearningboxplot}. The concrete modelling methodologies for DT-creation are denoted in Fig.\ref{fig:TopMMDescOverYear_ScatterBubble_Legend} as a matrix-like scatter plot over time. The size of the scatter point is based on the published research papers utilizing the corresponding methodology. \\
In addition, the individual extracted research items, including information on types, modelling task (MT), modelling methodology (MM), learning paradigm (LP), architecture (Arch) and experiments (Exp), are denoted in Tab.\ref{tab:research_summary}. For ensuring a compact table representation, the abbreviated terminology is used, e.g. data-based (D), hybrid (H), semi-supervised (SS), supervised (S), unsupervised (U), self-supervised (SFS), transfer learning (TL) and reinforcement learning (RL). The modelling methodologies are abbreviated as Autoencoder (AE), Long-Short-Term Memory (LSTM), Convolutional Neural Network (CNN), Physics-Informed Neural Network (PINN), Finite Element Analysis (FEA), Heat Transfer Equations (HTE), Decision Tree (DTR), Neural Network (NN), Attention Mechanism (AM), Bayesian Network (BN), Graph Neural Network (GNN), Gaussian Mixture Models (GMM), Support Vector Machine (SVM), Multi-Layer Perceptron (MLP), and Echo State Networks (ESN).

\newpage

\begin{table}[H]
\centering
\caption{Summary of primary studies. Modelling Task (MT), Modelling Methodology (MM), Learning Paradigm (LP), Arch (Architecture), Exp (Experiments).}
\label{tab:research_summary}
\begin{tabularx}{\textwidth}{>{\centering\arraybackslash\hsize=.6\hsize}X>{\centering\arraybackslash\hsize=1.0\hsize}X>{\centering\arraybackslash\hsize=1.4\hsize}X>{\centering\arraybackslash\hsize=0.6\hsize}X>{\centering\arraybackslash\hsize=.5\hsize}X>{\centering\arraybackslash\hsize=.5\hsize}X>{\centering\arraybackslash\hsize=1.4\hsize}X}
\toprule
\textbf{Type} & \textbf{MT} & \textbf{MM} & \textbf{LP} & \textbf{Arch} & \textbf{Exp} & \textbf{Author} \\
\midrule
D & Classification & AE, LSTM & SS & yes & yes & Lu et al.\cite{lu_semi-supervised_2022} \\
\midrule
D & Classification & NN, GA & SS & no & yes & Orošnjak et al.\cite{orosnjak_signals_2023} \\
\midrule
H & Regression, Classification & RNN & S, RL & yes & no & Jaensch et al.\cite{jaensch_digital_2018} \\
\midrule
D & Classification & CNN, AM & S & partly & yes & Li et al.\cite{li_novel_2023}\\
\midrule
D & Regression & AE & SFS, TL & no & yes & Yan et al.\cite{yan_toward_2023}\\
\midrule
H & Classification & CNN & S, U & yes & yes & Yu et al.\cite{yu_dynamic_2023}\\
\midrule
H & Regression & DTR, MLP, SVM, HTE, FEA & S, U, TL & partly & yes & Valdés et al.\cite{valdes_deterministic_2020} \\
\midrule
H & Regression & PINN & U & no & yes & Hosseini et al.\cite{hosseini_single-track_2023} \\
\midrule
H & Classification & FEA, DTR & S & partly & yes & Gawade et al.\cite{gawade_leveraging_2022} \\
\midrule
D & Classification & CNN & U, TL & - & yes & Sun et al.\cite{sun_fault_2024} \\
\midrule
H & Classification & CNN & S, TL & partly & yes & Jauhari et al.\cite{jauhari_modeling_2023} \\
\midrule
H & Regression & SSM, RNN, ESN & S & partly & partly & Dettori et al.\cite{dettori_optimizing_2022}\\
\midrule
D & Regression, Classification & DTR, SVC, NN & S & no & yes & Chen et al.\cite{chen_multisensor_2023} \\
\midrule
D & Clustering & AE & U & partly & yes & Cancemi et al.\cite{cancemi_unsupervised_2023} \\
\midrule
D & Classification & CNN, AE & S & partly & partly & Bauer et al.\cite{bauer_artificial_2022}\\
\midrule
D & Classification & BN & SS & partly & yes & Qi et al.\cite{qi_semi-supervised_2024} \\
\midrule
D & Classification & LSTM, AE & S, U & yes & yes & Hu et al.\cite{hu_digital_2023} \\
\midrule
H & Regression & GNN & S & no & yes & Hernandez et al.\cite{hernandez_thermodynamics-informed_2024} \\
\midrule
H & - & CNN, ANN & S, TL & yes & partly & Alexopoulos et al.\cite{alexopoulos_digital_2020}\\
\midrule
D & Classification, Regression & DTR, AM, LSTM & U & yes & yes & Zhang, Rui et al.\cite{zhang_research_2022} \\
\midrule
D & Classification & AE & SS & partly & yes & Castellani et al.\cite{castellani_real-world_2021} \\
\midrule
D & Classification, Clustering & DTR, NN & U, S & partly & yes & Naser\cite{naser_digital_2022} \\
\midrule
H & Classification & SVM & S & partly & yes & Gaikwad et al.\cite{gaikwad_toward_2020} \\
\midrule
D & Classification & GMM & U & yes & yes & Ladj et al.\cite{ladj_knowledge-based_2021} \\
\midrule
H & Clustering & FEA & SS, TL & yes & yes & Xia et al.\cite{xia_digital_2023} \\
\midrule
D & Classification & CNN & S & partly & yes & Tang et al.\cite{tang_particle_2023} \\
\midrule
H & Regression & RNN, CNN, LSTM, AE & U & partly & yes & Gupta et al.\cite{gupta_three-dimensional_2022} \\
\midrule
H & Classification & CNN & S & no & yes & Parola et al.\cite{parola_convolutional_2023} \\
\midrule
D & Regression & MLP, SVR, DTR & S & yes & yes & Boukredera et al.\cite{boukredera_enhancing_2023} \\
\midrule
D & Regression & MLP, NN & S & partly & yes & Schroer et al.\cite{schroer_feature_2024} \\
\midrule
H & - & - & - & yes & no & Abburu et al.\cite{abburu_cognitwin_2020} \\
\bottomrule
\end{tabularx}
\end{table}

\newpage
\section{Evaluating the Research Questions}
\label{sec:evaluatereview}
This critical analysis of the primary studies aims to address the research questions outlined earlier. Each question is explored through the lens of the findings from these studies. This section summarizes the findings and synthesizes the evidence to provide a coherent understanding of the current landscape and future directions in digital twin technologies for industrial manufacturing.

\subsubsection{Research Question 1:} \emph{What are the state-of-the-art modelling methodologies, and how has their usage evolved? Are Encoder-Decoder architectures and attention mechanisms on the rise, possibly signalling the adoption of transformer-like architectures to the industrial DT domain?}

\noindent The used modelling methodologies as the basis for DT-creation in the context of process industry are Convolutional Neural Network (CNN) e.g. \cite{li_novel_2023},\cite{sun_fault_2024},\cite{bauer_artificial_2022},\cite{parola_convolutional_2023},  followed by Autoencoder (AE), e.g. \cite{lu_semi-supervised_2022}, \cite{cancemi_unsupervised_2023},\cite{castellani_real-world_2021},\cite{yan_toward_2023}. Lue et al. \cite{lu_semi-supervised_2022} uses LSTMs in combination with Autoencoder for condition monitoring of magnesium furnace processes. The work combines data on electrical currents with segmented image features of the furnace flames. Li et al. \cite{li_novel_2023} uses a customized CNN with deconvolution and convolution layers in combination with an attention layer on top to detect anomalies in the SWaT and WADI datasets. We have yet to see the adoption of state-of-the-art modelling methodologies of Large Language Models for the industrial DT-creation context. A notable exception in this context is Yan et al. \cite{yan_toward_2023}, who uses a masked autoencoder that pre-trains in a self-supervised way using three novel pre-text tasks (inverse forecasting, coarser forecasting and anomaly forecasting). Results indicate robust performance and "[...] may pave the way for leveraging pre-training approaches for multivariate time-series forecasting in the context of digital twins"; however, it is crucial to understand the behaviour of such pre-training strategies and the influence on various DT-related tasks \cite{yan_toward_2023}.

\subsubsection{Research Question 2:} \emph{What are the commonly utilized learning paradigms (i.e. unsupervised, supervised, semi-supervised, self-supervised or transfer-learning)? What is the distribution of data-driven and hybrid DT-modelling approaches? Is the self-supervised learning paradigm for generating general and transferable knowledge for various DT-related tasks already explored in the industrial DT domain?}

\noindent Most identified research items in this study use a supervised learning paradigm (e.g. \cite{hernandez_thermodynamics-informed_2024,alexopoulos_digital_2020,gaikwad_toward_2020}) to model the DT's use-case (see. Fig.\ref{fig:tasktypelearningboxplot}). Unsupervised learning paradigms are commonly used in anomaly or fault detection (\cite{cancemi_unsupervised_2023,ladj_knowledge-based_2021}). Ladj et al. use a Gaussian Mixture Model to determine the classification thresholds and also integrate a "data-knowledge closed loop system which combines a data-driven approach with a knowledge-driven approach, where expert rules are extracted and inferred in order to interpret and augment the results of data processing" \cite{ladj_knowledge-based_2021}. The self-supervised learning paradigm, in combination with transferring knowledge to other DT-related tasks (e.g. prediction, simulation, anomaly detection), is not well-explored in the domain of DTs (notable exception is \cite{yan_toward_2023}). We think there is substantial potential in such learning paradigms, due to the vast availability of multi-modal, high-dimensional and high-volume IIoT-data, the abundance of labelled data, and the diverse expectations and use-case scenarios of DTs in industrial manufacturing. Around 40\% of the analyzed primary studies employ a hybrid modelling approach, combining data-driven and physics-based modelling to tackle problems like small amounts of (labelled) data, or accuracy. Notable hybrid-based models are Hosseini et al. \cite{hosseini_single-track_2023}, who use physics-informed neural networks (PINNs) as a simulator for temperature profiles of laser powder bed fusion (LPBF) processes w.r.t. the different input process parameters and material thermal properties.

\subsubsection{Research Question 3:} \emph{Are DT research and application studies more focused on evaluation on classification, clustering tasks (e.g. anomaly detection) or regression tasks (e.g. forecasting or imputation)?}

\noindent Given the conducted review results, it is evident that most research in the selected primary studies targets classification or clustering tasks, which are common for DT-related tasks like anomaly detection or failure identification. etc. (\cite{cancemi_unsupervised_2023,ladj_knowledge-based_2021,li_novel_2023,yu_dynamic_2023}). Less well represented are scientific publications in the context of DT in the process industry for regression tasks like forecasting or imputation (\cite{boukredera_enhancing_2023,schroer_feature_2024,hernandez_thermodynamics-informed_2024}). Hernandez et al. propose a method based on graph neural networks and encoder-decoder structures to predict the time evolution of an arbitrary dynamical system using both geometric and thermodynamic inductive biases; however, this work is limited to in-silico experiments, i.e. DT application of this methodology for industrial processes is still missing \cite{hernandez_thermodynamics-informed_2024}.

\section{Discussion and Conclusion}
\label{sec:discussionandconclusion}
In this study, we conduct a structured literature review to unravel the modelling methodologies, learning paradigms, and task-related evaluation aspects of Digital Twins in the context of the process industry. We first state the research questions, then formulate the search query and perform the search across various literature databases. We subsequently filter the resulting list based on includance and excludance rules and evaluate the full texts of the reduced list based on defined quality criteria to capture the relevancy of the analyzed papers for answering the specific research questions. The selected 31 primary papers apply DT-related modelling methodologies to various use cases in the industrial manufacturing domain. We synthesize the findings visually and textually to answer the research questions. 
We observe abundant research for self-supervised and transfer-learning paradigms (e.g. transfer knowledge to other DT-related tasks like prediction, simulation or anomaly detection). We have identified a research gap and think there is substantial potential in such learning paradigms due to the vast availability of multi-modal, high-dimensional and high-volume IIoT-data, the abundance of labelled data, and the diverse expectations and use-case scenarios of DTs in the industrial manufacturing domain.

\begin{credits}
\subsubsection{\ackname} This work received funding as part of the Trineflex project (trineflex.eu), which has received funding from the
European Union’s Horizon Europe research and innovation programme under
Grant Agreement No 101058174. Funded by the European
Union. Views and opinions expressed are however those of
the author(s) only and do not necessarily reflect those of
the European Union, the Austrian federal government or the
federal state of Upper Austria. Neither the European Union
nor the granting authority can be held responsible for them.
\subsubsection{\discintname}
The authors have no competing interests to declare that are
relevant to the content of this article.
\end{credits}
\bibliographystyle{splncs04}
\bibliography{references}

\end{document}